\documentclass[12pt, anon, nosubfloats]{l4dc2024}

\usepackage[utf8]{inputenc} 
\usepackage[T1]{fontenc}    
\usepackage{hyperref}       
\usepackage{url}            
\usepackage{booktabs}       
\usepackage{amsfonts}       
\usepackage{nicefrac}       
\usepackage{microtype}      
\usepackage{xcolor}         
\usepackage{bbm}
\usepackage{graphicx}
\usepackage{subfig}
\usepackage{enumitem}

\definecolor{burntorange}{rgb}{0.8, 0.33, 0.0}

\newcommand{\xxnote}[3]{}
\ifx\hidenotes\undefined
  \renewcommand{\xxnote}[3]{\color{#2}{#1: #3}}
\fi

\graphicspath{{figs/}}
\title{Real-World Fluid Directed Rigid Body Control via Deep Reinforcement Learning}

\author{\Name{Mohak Bhardwaj}\thanks{Google DeepMind}\thanks{University of Washington}\hfill  
  \Name{Thomas Lampe}\footnotemark[1]\hfill 
  \Name{Michael Neunert}\footnotemark[1] \\ 
  \Name{Francesco Romano}\footnotemark[1] \hfill
  \Name{Abbas Abdolmaleki}\footnotemark[1]\hfill 
  \Name{Arunkumar Byravan}\footnotemark[1]\\
  \Name{Markus Wulfmeier}\footnotemark[1] \hfill
  \Name{Martin Riedmiller}\footnotemark[1] \hfill 
  \Name{Jonas Buchli}\footnotemark[1]
  \\
  }

\begin{document}

\maketitle

\begin{abstract}
    Recent advances in real-world applications of reinforcement learning (RL) have relied on the ability to accurately simulate systems at scale. 
    However, domains such as fluid dynamical systems exhibit complex dynamic phenomena that are hard to simulate at high integration rates, limiting the direct application of modern deep RL algorithms to often expensive or safety critical hardware. In this work, we introduce "Box o’ Flows", a novel benchtop experimental control system for systematically evaluating RL algorithms in dynamic real-world scenarios. 
    We describe the key components of the Box o' Flows, and 
    through a series of experiments demonstrate how state-of-the-art model-free RL algorithms can synthesize a variety of complex behaviors via simple reward specifications.
    Furthermore, we explore the role of offline RL in data-efficient hypothesis testing by reusing
    past experiences. We believe that the insights gained from this preliminary study and the availability of systems like the Box o' Flows support the way forward for developing systematic RL algorithms that can be generally applied to complex, dynamical systems. Supplementary material and videos of experiments are available at \url{https://sites.google.com/view/box-o-flows/home}.
\end{abstract}

\section{Introduction}\label{sec:introduction}
Reinforcement learning promises to deliver a principled, general-purpose framework for generating control policies for complex dynamical systems directly from experiential data, without the need for domain expertise~\citep{sutton2018reinforcement}. Indeed, modern \textit{deep} RL approaches that leverage expressive neural networks 
for function approximation 
have led to breakthroughs in a variety of domains, such as  
game-playing~\citep{mnih2013playing, schrittwieser2020mastering, mnih2015human}, 
protein folding~\citep{jumper2021highly}, control of tokamak plasmas in nuclear fusion reactors~\citep{degrave2022magnetic}, and real-world robotics~\citep{tan2018sim, handa2022dextreme}. \looseness=-1

However, a key ingredient in the success of these applications has been the ability to accurately simulate these systems at scale, and constructing such simulation environments themselves requires significant 
human effort and knowledge, thus forgoing the original promise of removing the need for domain expertise. For instance, leading approaches for learning-based locomotion and dexterous manipulation~\citep{tan2018sim, kumar2021rma, fu2021minimizing, handa2022dextreme, pinto2017asymmetric} rely on a \textit{sim-to-real} paradigm to learn robust policies in simulation that can be directly transferred to the real world.
Even when policies are learned directly on real hardware, practitioners often rely on simulation to gain intuition about the problem domain, and make critical design decisions such as the choice of algorithm, reward functions and other hyperparameters~\citep{pmlr-v164-lee22b, DBLP:conf/rss/SchwabSMNLAHHNR19}.

In addition to human expertise involved in simulation design, 
the high sample complexity of current RL algorithms necessitates fast simulations to achieve reasonable wall clock times for training. While this is possible for domains such as video games and rigid-body systems~\citep{todorov2012mujoco, liang2018gpuaccelerated}, for several real-world problems satisfying this need becomes increasingly expensive or outright impossible. Examples 
include systems involving non-steady fluid dynamics and/or continuum mechanics (e.g. flying, swimming, soft matter based mechatronic systems), and
multi-scale problems 
that occur
in biological systems or digital twins of large industrial systems. How can we scale RL to such systems?
This work focuses on one such domain - the control of coupled mechanical-fluid dynamic systems. Here, the fact that one can not assume steady state dynamics hugely increases the complexity of simulations.
For example, consider an Unmanned Aerial Vehicle operating in off-nominal regimes such as high angle of attack or ground/obstacle effects. Here, the turbulent air flows that are generated can be difficult to model, and create instabilities that nominal controllers are incapable of handling. 
While there is a growing literature on learning control policies in the presence of non-steady fluid flows that utilize
simulation~\citep{verma18swim}, and the dynamics are known in principle, simulating them requires supercomputers which is beyond the resources of most practitioners. 
The study of such systems raises interesting questions that have several implications for real-world deployment of reinforcement learning. 
\begin{enumerate}
\item How do we design experiments to characterize the capabilities of a system that is hard to simulate at scale? 
\item How do we ensure sample efficient learning given limited data collection rates?
\item How can we efficiently re-use prior experience to test different hypotheses, and aid the learning of new behaviors?
\end{enumerate}

\begin{figure}
    \centering
    \includegraphics[trim={0 10cm 0 10cm}, clip, width=\textwidth]{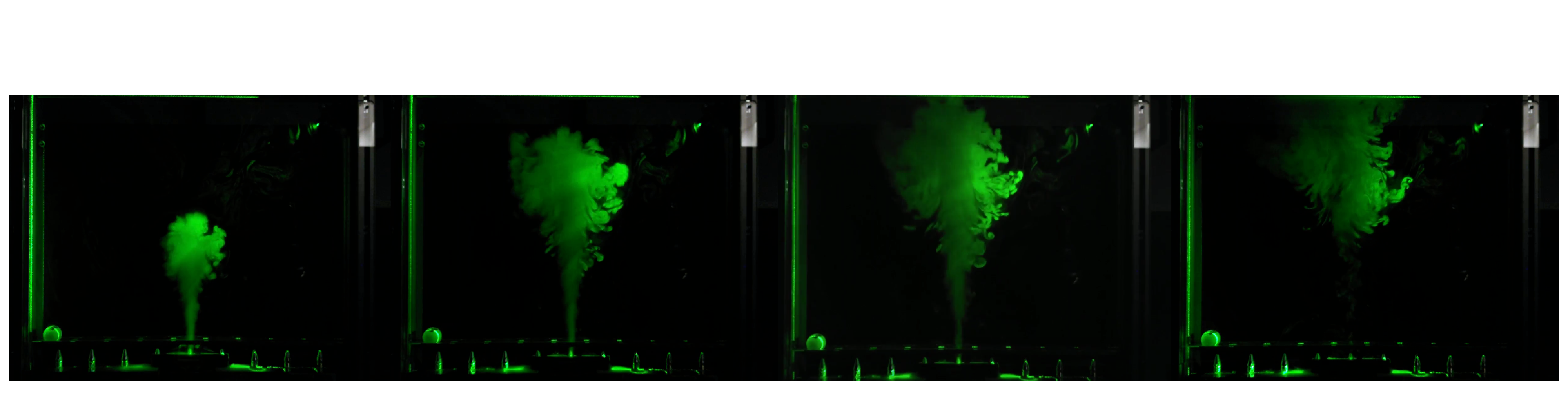}
    \caption{Smoke visualizes the complex flow field that emerges from a single valve with constant flow. This illustrates the complex relationship between actuator and the flow field and ultimately its effects on the balls. This relationship is further complicated when several actuators are acting simultaneously.}
    \label{fig:smoke_generator}
\end{figure}

To investigate these questions, we have developed a novel fluid-dynamic control system dubbed "Box o' Flows". This system consists of 9 upward facing nozzles arranged in parallel with a proportional pneumatic valve per nozzle regulating the airflow. The valves can be controlled programmatically to create complex pressure fields between two parallel panels forming a box. The airflow can be used to control the state of rigid objects, such as colored balls, that are placed inside. The setup is also equipped with an RGB camera capturing the box and objects inside it~(Fig.~\ref{fig:box_o_flows} provides a detailed overview).
The system is intentionally designed to be impossible to simulate accurately at the high integration rates required by deep RL algorithms, 
and exhibits complex non-steady fluid dynamics which makes (unknowingly) injecting prior human knowledge, or hand-designing control policies hard in practice. 
In Fig.~\ref{fig:smoke_generator} we demonstrate fluid patterns generated by the 
air flowing through the nozzles. 

This work serves as a preliminary investigation of how model-free RL can be used to learn a variety of dynamic control tasks on the Box o' Flows directly in the real world, as well as characterize hardware capabilities. 
We limit the algorithms tested to the state-of-the-art Maximum A-posteriori Policy Optimization (MPO)~\citep{abdolmaleki2018maximum}, with fixed hyperparameters across different experiments. Desired behaviors are described via minimally specified rewards functions, which gives the RL agent the freedom to find interesting control strategies. Furthermore, we test how offline RL can be used as a means for hypotheses testing by training new policies on logged data from past experiments, and intermittently evaluating them on the real system. Our framework can generate diverse dynamic behaviors to control the state of multiple rigid objects (table tennis balls) such as hovering, rearrangement, stacking and goal-reaching (detailed in Sec.~\ref{sec:experiments}). In summary, our main contributions are:

\begin{figure}
    \centering
    \includegraphics[width=0.3\textwidth]{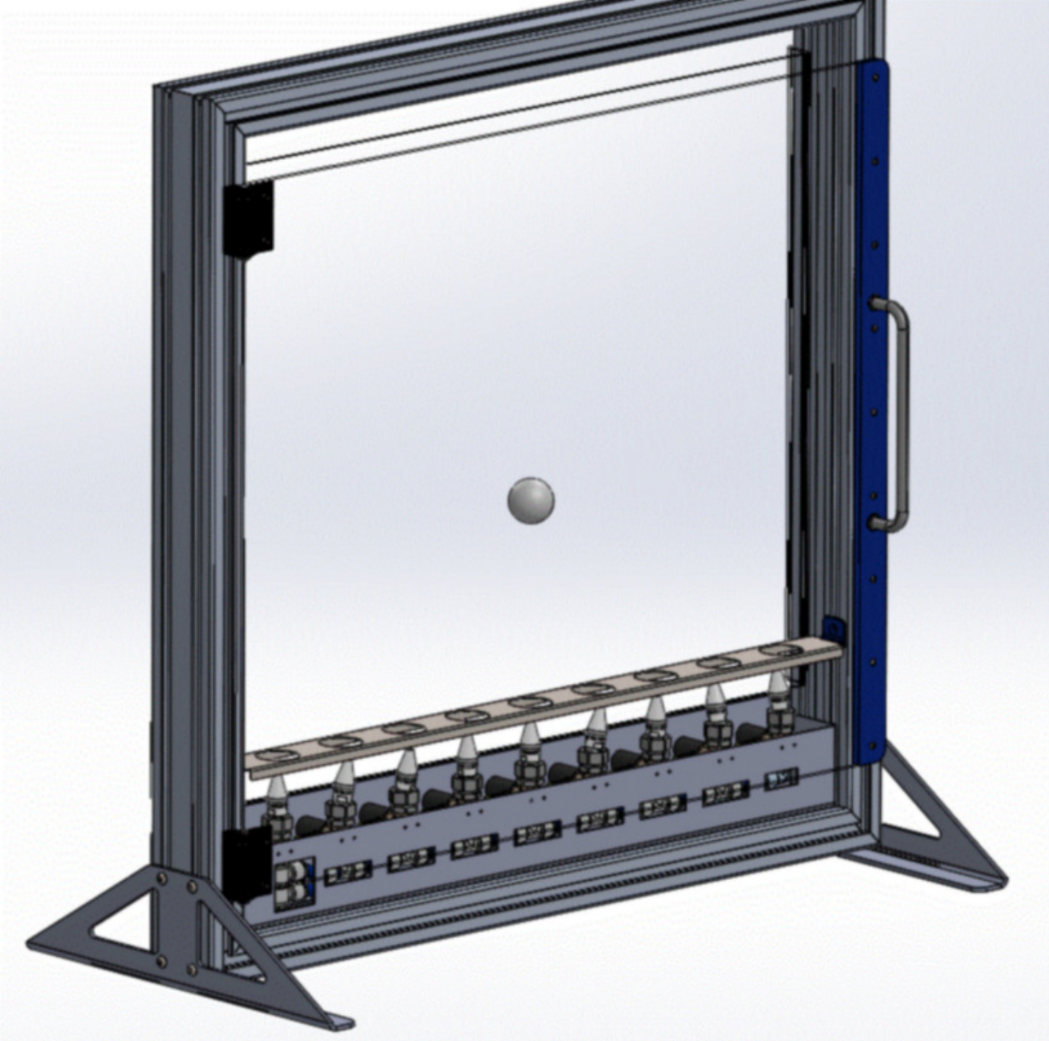}
    \includegraphics[width=0.5\textwidth]{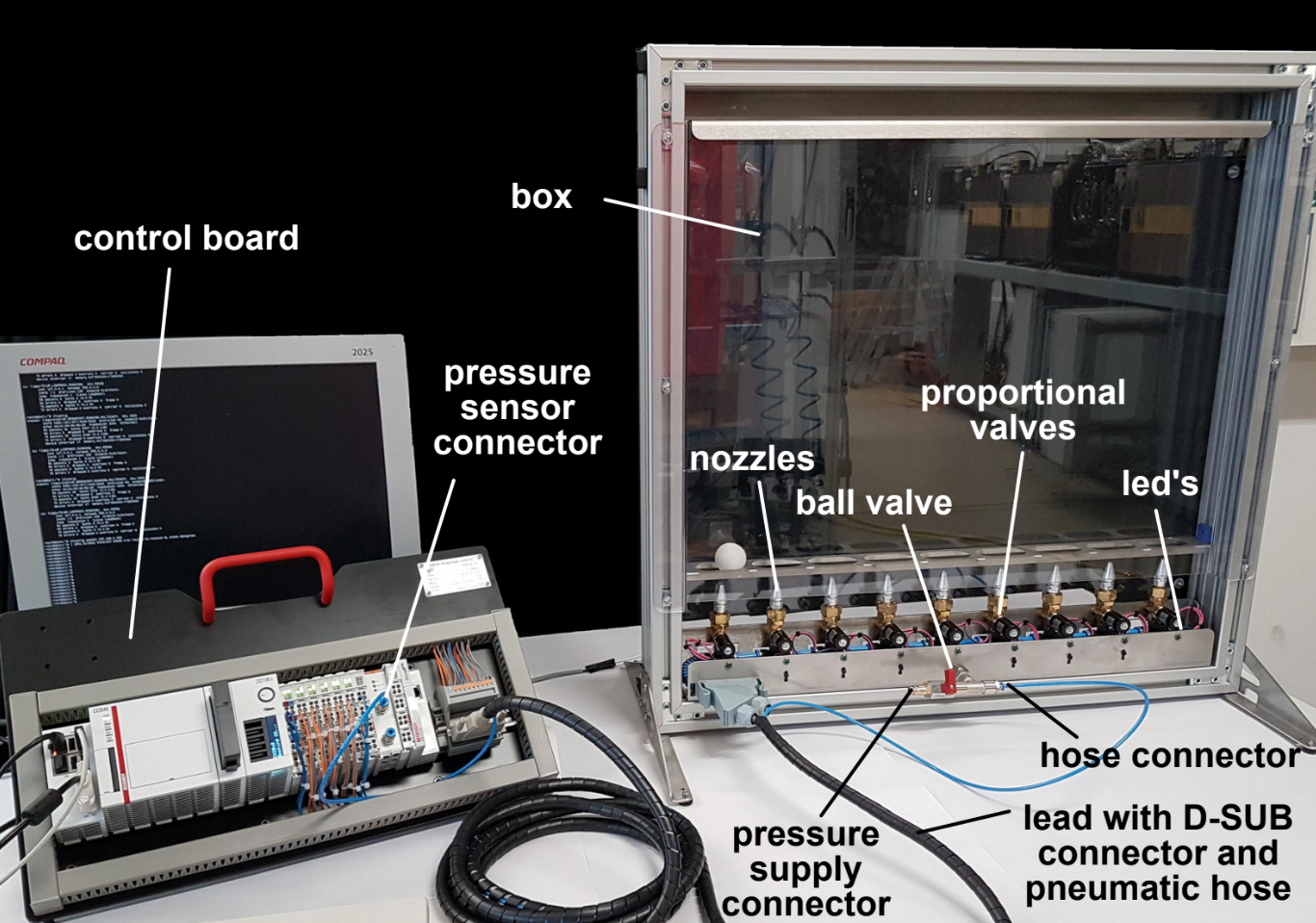}
    \caption{An overview of the different components of bench-top Box o' Flows system.}
    \label{fig:box_o_flows}
\end{figure}
\begin{itemize}[wide, labelwidth=!, itemindent=!,labelindent=0pt]
    \item We present a novel benchtop fluid-dynamic control system geared towards real-world RL research.
    \item We demonstrate the application of sample-efficient, model-free RL to learning dynamic behaviors and analyzing hardware capabilities.
    \item We explore how offline RL with past data can be used to test various hypotheses when simulation is not available.
\end{itemize}

\section{Box o' Flows - System Overview}\label{sec:system_overview}
In this section we describe the Box o' Flows system as shown in Fig.~\ref{fig:box_o_flows}. The system comprises of a 70cmX70cm square aluminum frame on which a black opaque back panel and a transparent front panel are mounted, creating a shallow box of roughly 60mm depth. Mounted at the bottom edge of this box is a blade consisting of 9 proportional flow control valves (SMC PVQ 30), each attached to a nozzle facing upwards. An LED strip is mounted on the remaining three sides to evenly illuminate the interior of the box. Objects, such as the colored table tennis balls used in this work, can be placed within the space inside the box, so that their state can be controlled via the airflow.

All valves share a common air supply that is hooked up to an air pump and fed via the proportional control valves at 6 bar. By connecting all the nozzles to a single pump, the supply pressure and consequently the flow across the nozzles drops when multiple valves are opened simultaneously. This cross coupling has been added intentionally, to increase the complexity of the system behaviour. Further, the system can only measure the overall supply pressure and not the pressure or flow at each valve.

Communication with the valves and sensors is realized through EtherCAT, a realtime ethernet protocol providing synchronization between the individual nozzles. The control system runs on an intel-i7 based Beckhoff industrial PC running Linux and the EtherLab EtherCAT master~\citep{etherlab}. A machine vision camera (BASLER Ace acA1920-40gc) is attached via GigE Ethernet and captures RGB images of the interior of the box. While the underlying Ethercat bus runs at higher rates, for the experiments described here a control rate of 20 Hz has been used. 

\begin{figure}[t!]
    \vspace{-10mm}
    \centering
    \subfloat[][]{\includegraphics[trim={0cm 0cm 1cm 1cm}, clip, width=0.3\linewidth]{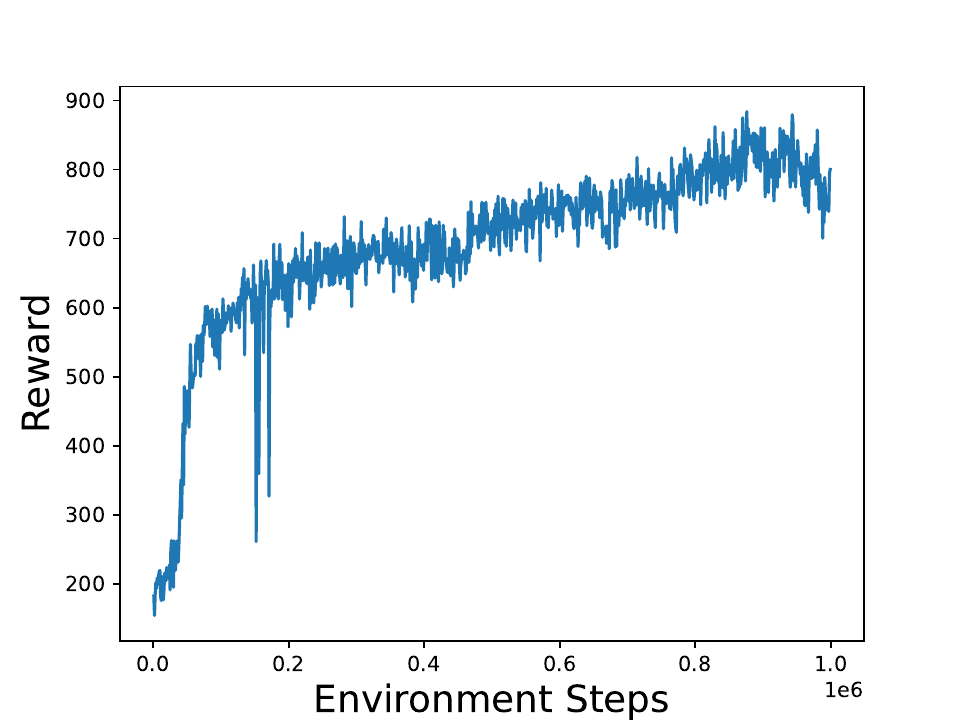}}\hfill
    \subfloat[][]{\includegraphics[trim={5cm 9cm 3cm 1cm}, clip, width=0.7\linewidth]{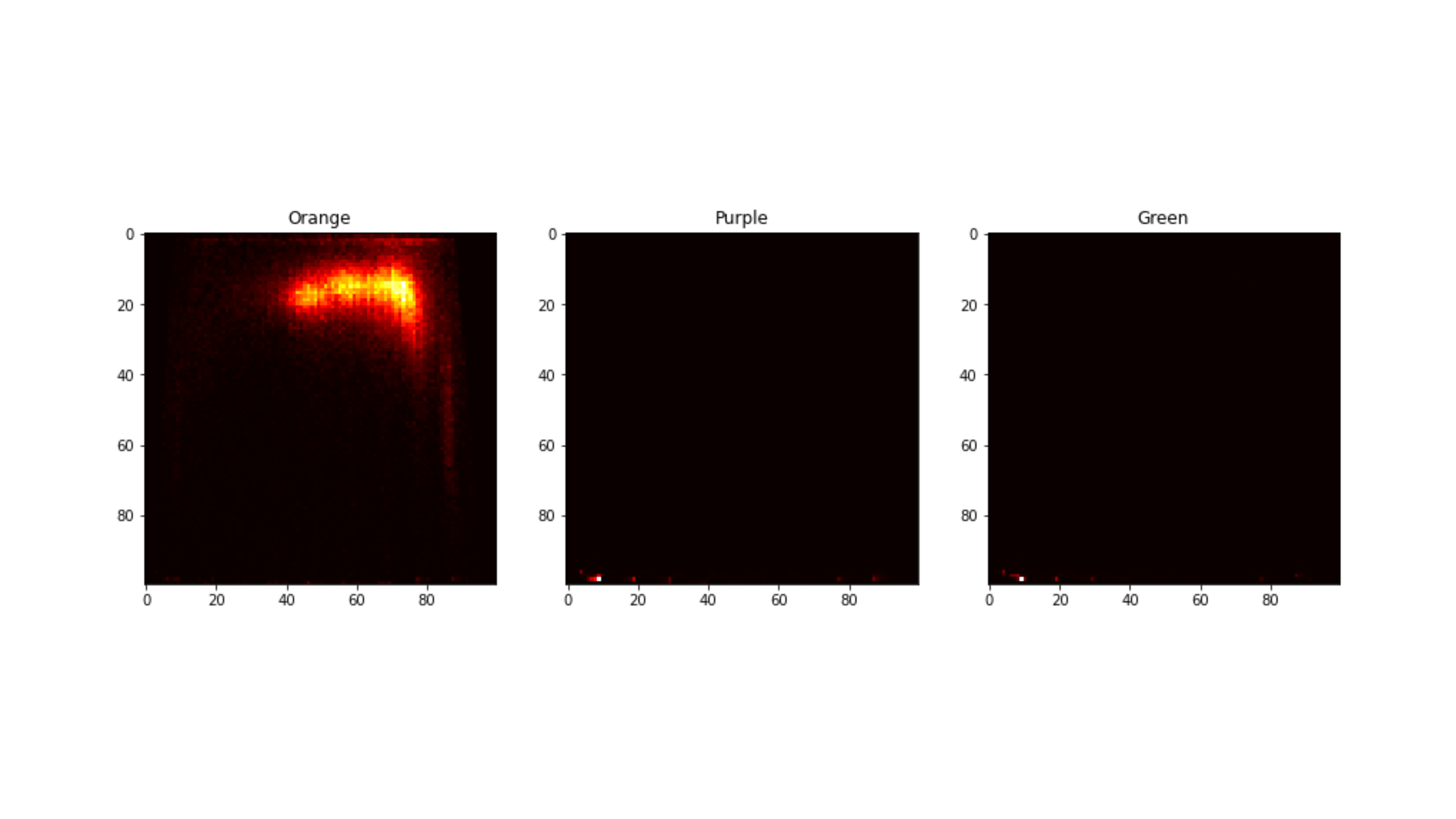}}\\[-0.5ex]
    \subfloat[][]{
    \centering
    \includegraphics[trim={0cm 10cm 0 10cm}, clip, width=0.9\linewidth]{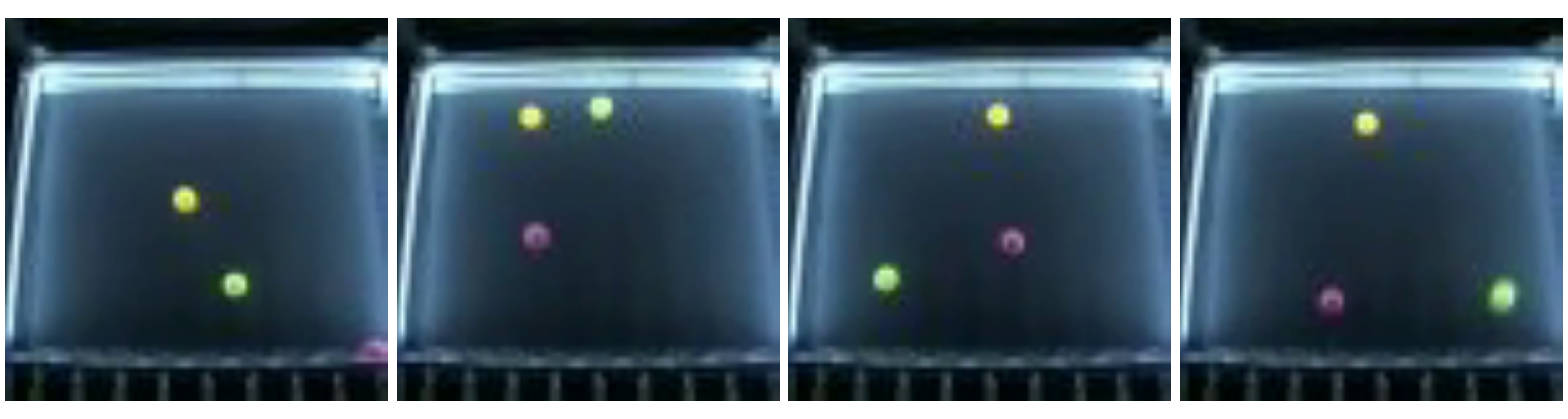}}\hfill
    \caption{Reinforcement learning applied to the task of maximizing the height of orange ball in presence of distractors (purple and green). The non-steady fluid dynamics of interacting objects and complex actuator coupling makes it hard to hand-design controllers. (a) Reward curve (b) Heatmap visualization of states visited by learned policy (averaged over 100 episodes) (c) Filmstrip of an episode (More details in Sec.~\ref{sec:experiments})} 
    \label{fig:maximize_orange}
\end{figure}

\subsection{What Makes Box o' Flows a Hard Problem?}
\label{subsec:hardness_of_bof}
The Box o' Flows brings to light several key challenges in controlling real-world systems with complex dynamics. As a motivating example, consider a simple setting with three colored balls placed inside the box, and one must design a control policy to maximize the height of one of the balls, with the others being distractors, i.e their motion is not constrained.
(For reference, Fig.~\ref{fig:maximize_orange}(c) shows behavior learned by our framework).
While intuitively it may seem straightforward to hand-design a controller (eg. maximally open all valves), the nature of the Box o' Flows makes it hard in practice.

First, the cross coupling between actuators due to shared air supply means that maximally opening all valves will not work for this task since the pressure per valve will drop. This relation is also hard to model and changes unpredictably over time due to practical issues such as oil accumulation. Second, in the Box o' Flows there is a less direct relationship from the actuator space to the state space than a standard robotic system. The non-steady dynamics of the emerging flow given an actuation input is highly complex and stochastic, especially as the objects interact with each other, and the controller must account for this. Moreover, current methods for accurately simulating non-steady flows require large amounts of compute which precludes techniques like \textit{sim-to-real} RL that rely on cheap simulated data. \looseness=-1

Third, the system is highly under-observed as we can not directly measure the flow field inside the box, but only the supply pressure. One can only attempt to recover this information from a history of images of object motion from the camera stream. Finally, real-world data collection is a limiting factor. The current setup can collect approximately 1M environment steps per day, thus, experiments must be designed carefully for efficient data use. \looseness=-1

From the above, it is clear that hand-designing controllers is non-trivial even in simple settings, and model-based techniques that rely on accurate system identification or simulation can be prohibitively expensive. It is therefore more promising to consider efficient data-driven approaches that can overcome these constraints.

\section{Methods}\label{sec:methods}
We focus on sample-efficient, model-free RL algorithms that can facilitate learning control policies from limited real-world experience, both via online interactions and offline datasets.
To this end, we leverage a high performance off policy actor-critic algorithm, Maximum Aposteriori Policy Optimization (MPO)~\citep{abdolmaleki2018relative, abdolmaleki2018maximum}.
At iteration $k$, MPO updates the parameters $\phi$ and $\theta$ of the critic $Q^{\pi^{k}}_{\phi}$ and policy $\pi^{k}_{\theta}(\cdot | s)$ respectively by optimizing \looseness=-1

\begin{equation}
    \label{eq:mpo_critic_loss}
    \min_{\phi} \left(r_t + \gamma Q^{\pi^{k-1}}_{\phi'}(s_{t+1}, a_{t+1} \sim \pi^{k-1}) - Q_{\phi}^{\pi^{k}}\left(s_t, a_t\right) \right)
\end{equation}

\begin{equation}
    \label{eq:mpo_policy_loss}
    \pi^{k+1}_{\theta} = \arg\min \mathrm{E}_{\mu}\left[KL(q(a|s) || \pi_{\theta}((a|s)))\right]
\end{equation}
where $q(a|s) \propto \exp(Q_{\phi}^{k}(s,a) \mu /\beta))$ is a non-parametric estimate of the optimal policy given a temperature $\beta$, and $KL\left(q(\cdot|s)||\pi(\cdot|s)\right)$ is the KL divergence, and $\mu$ is the distribution of states stored in a replay buffer. 
The efficient off-policy updates enable
MPO to demonstrate sample-efficient learning in high dimensional continuous control tasks. 
We refer the reader to~\cite{abdolmaleki2018relative} for a detailed derivation of the update rules. 
\paragraph{Offline RL:}
Since Box o' Flows is distinct from existing robotic setups, it can be a priori unknown what reward functions can lead to desired behaviors with online RL. This problem is aggravated by the lack of simulation and constrained data collection rates. Thus, it is vital to be able to to re-use prior experience to test hypotheses about new rewards. To this end, we focus on the offline RL paradigm that enables learning effective policies from logged datasets without further exploration~\citep{levine2020offline}. To deal with limited data coverage, modern offline RL algorithms~\citep{kumar2020conservative, cheng2022adversarially} rely on a concept of pessimism under uncertainty by optimizing performance lower bounds, such that the agent is penalized for choosing actions outside the data support. \looseness=-1

The actor update of MPO can be easily adapted to the offline setting. Given a dataset of transitions $\mathcal{D} = \lbrace\left(s_i, a_i r_i, s_{i+1}\right)\rbrace_{i=1}^{N}$ collected by a behavior policy $\mu_{B}$, we can modify the distribution of states in  Eq.~\ref{eq:mpo_policy_loss} from $\mu$ to $\mu_{B}$ (state distribution in $\mathcal{D}$) and non-parametric optimal policy to $q(a|s) \propto \exp(Q_{\phi}^{k}(s,a)\mu_{B}/\beta)$. The actor update thus encourages reward maximization while staying close to $\mu_B$. This is a special case of Critic Regularized Regression (CRR)~\citep{wang2020critic}, a state-of-the-art offline RL algorithm, and can be implemented it in a common framework with MPO. In our setting, we re-label data from prior online RL experiments with new rewards (in line with \citep{Davchev2021WishYW, Yarats2022DontCT, Lambert2022TheCO, Tirumala2023ReplayAE}), and train a CRR agent offline that is tested intermittently on the real system to validate policy performance. The minimal use of hardware enables us to test multiple policies instead of just one that continuously trains online. We now present our main empirical results. \looseness=-1
  
\section{Experiments}\label{sec:experiments}
We use a suite of dynamic control tasks to test the efficacy of our RL framework and study the physical capabilities of the Box o' Flows system. 

\paragraph{Setup:}
To delineate the interplay between hardware capabilities and algorithm performance, we keep our RL agent (Sec.~\ref{sec:methods}) fixed across all tasks. We use a distributed learning framework akin to~\cite{hoffman2020acme}, and select hyperparameters using a candidate task where optimal behavior is qualitatively known (see below). The actor and critic are represented by feedforward neural networks, and object state by a history of pixel xy coordinates measured from the vision system via a blob detector. The 9-dim action space represents degree of valve opening in the range $\left[0,1\right]$. Object locations are reset using random air bursts at the beginning of every episode (1000 steps long at 20Hz).We describe desired behaviors via simple rewards based on desired object configurations, which gives the RL agent the freedom to find interesting control strategies. Next, we describe the tasks in detail.\footnote{A complete description of rewards and hyperparameters can be found in the supplementary material at \url{https://sites.google.com/view/box-o-flows/home}} \looseness=-1

\subsection{Learning Dynamic Behaviors with Online RL }\label{subsec:online_rl}
\paragraph{Hovering with Distractors:}\label{subsec:maximize_orange}
We first consider the task of maximizing the height of a target ball (orange) in the presence of distractors (purple and green), and use it to select relevant hyperparameters. Intuitively, a near-optimal strategy is to place the distractors near a bottom corner and use other valves to hover the target ball. However, as described in Sec.~\ref{subsec:hardness_of_bof}, complex actuator coupling and non-steady flow patterns make it hard to hand-design such a controller. 
We test whether our MPO agent can recover this intuitive policy, by training it using a reward proportional to the pixel y coordinate of only the target ball, normalized to $[0.0,1.0]$ (based on maximum and minimum coordinate values). Fig.~\ref{fig:maximize_orange}(a) presents the reward obtained over environment steps during training that shows the agent is able to obtain near-optimal reward in about 1M steps. In Fig.~\ref{fig:maximize_orange}(b), we visualize the learned behavior via coarsely discretized heatmaps of ball locations over the last 100 training episodes, which show that the agent successfully learns the intuitive policy of maintaining the  target ball near the top while pushing the distactors near the bottom left. 

\begin{figure}[t!]
    \vspace{-15mm}
    \centering
    \subfloat[][]{\includegraphics[trim={0cm 0cm 1cm 1cm}, clip, width=0.3\linewidth]{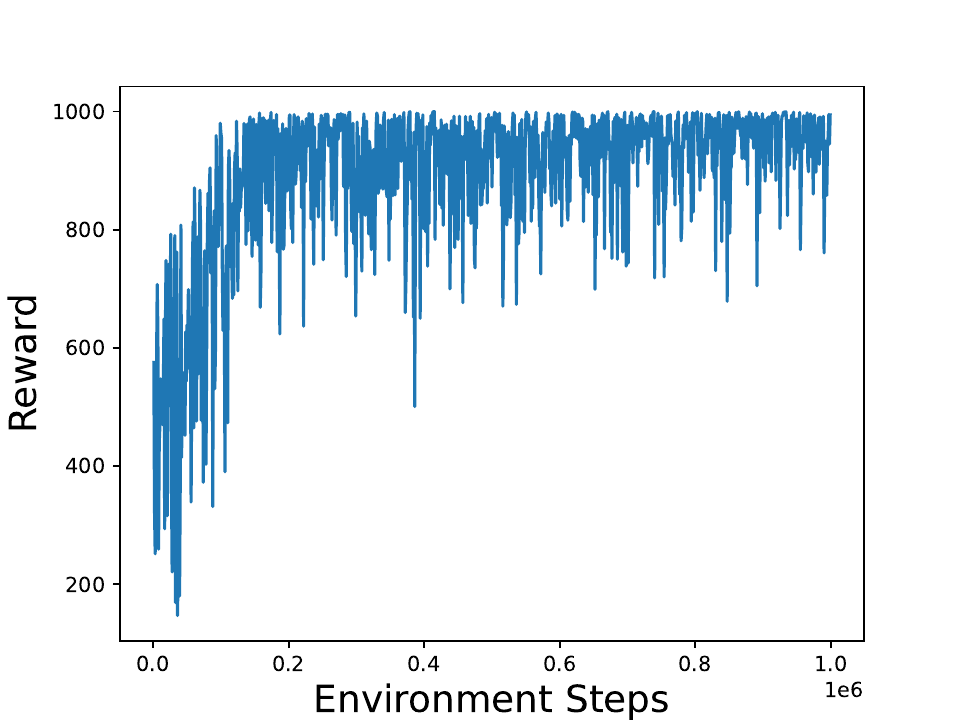}}\hfill
    \subfloat[][]{\includegraphics[trim={5cm 9cm 3cm 1cm}, clip, width=0.7\linewidth]{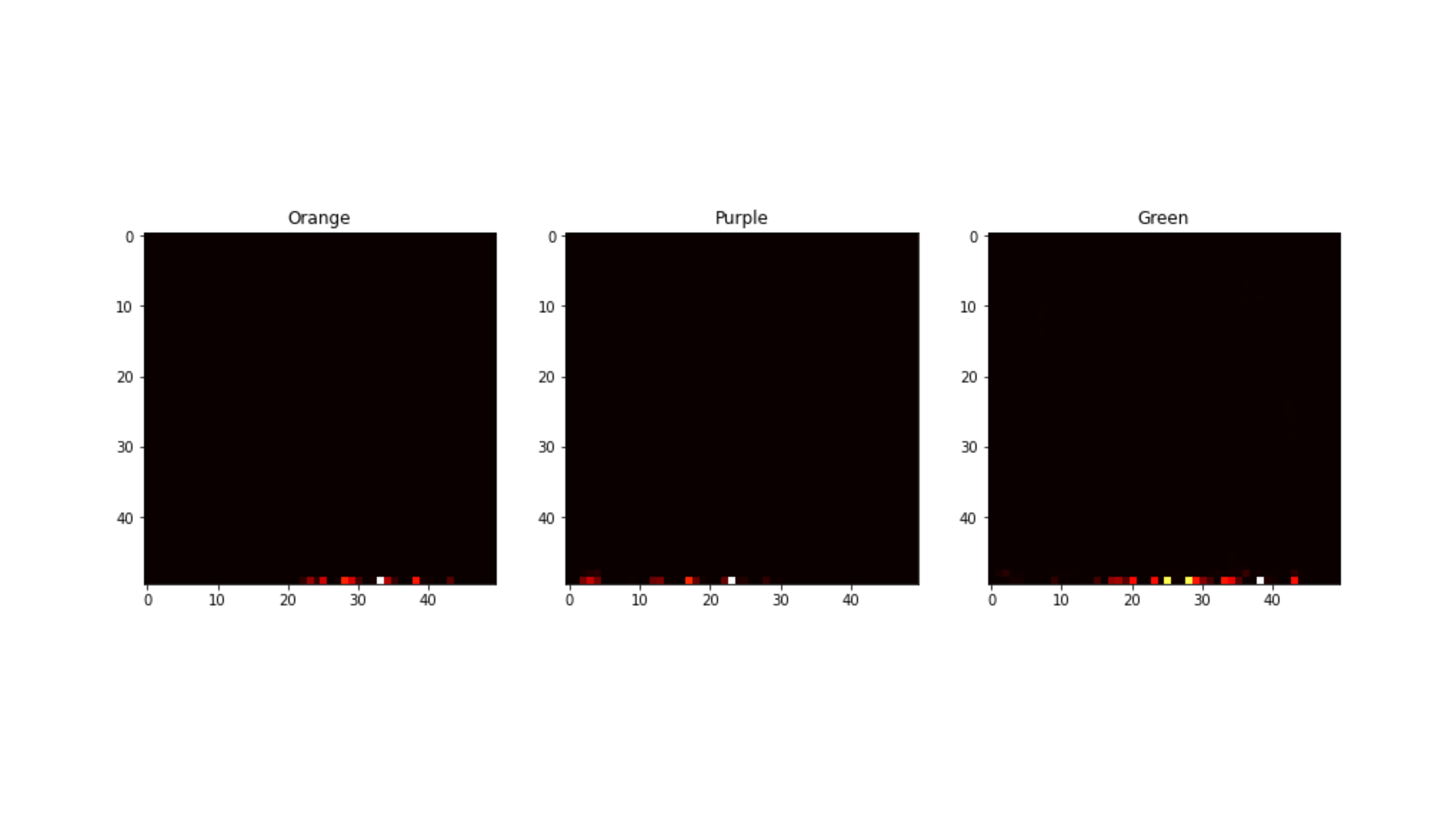}}\\[-0.5ex]
    \subfloat[][]{
    \centering
    \includegraphics[trim={0cm 9cm 0 9cm}, clip, width=0.9\linewidth]{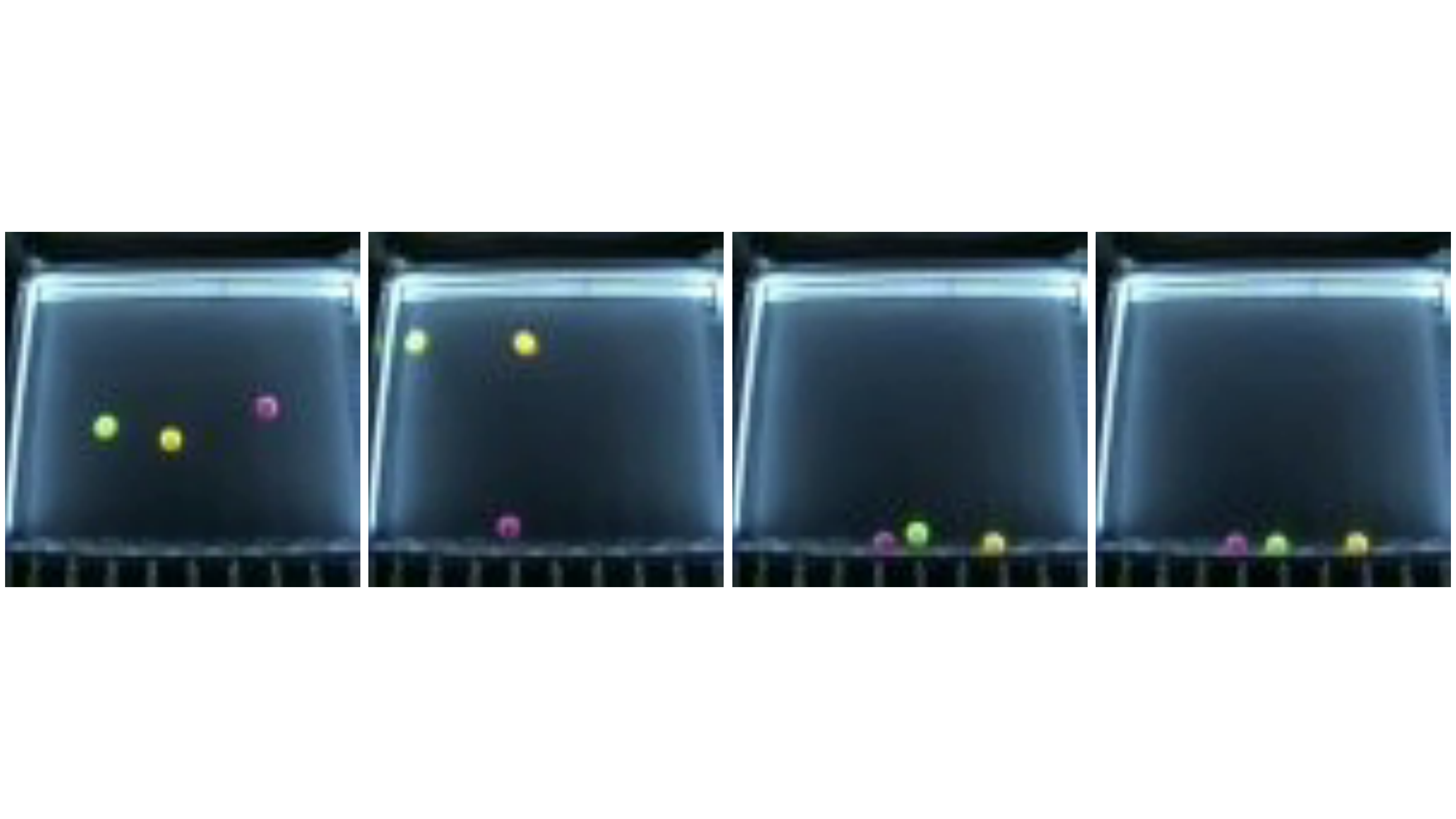}}\hfill
    \caption{Task: Orange in right, purple in left (a) Reward curve and (b) Heatmap visualization of states visited by learned policy (averaged over 100 episodes) (c) Filmstrip of an episode.}
    \label{fig:orange_right_purple_left}
\end{figure}
\paragraph{Object Rearrangement:}
Next, we consider a harder task where the agent must place two target balls (orange and purple) anywhere in the right and left halves of the box respectively, with the green ball being a distractor. Here, it is hard to even intuitively reason about optimal behavior as it depends on the initial object locations which are randomized.
We provide our agent a sparse reward equal to the product of the horizontal distances from the respective goal regions, which forces it to accomplish both tasks. As shown in Fig.~\ref{fig:orange_right_purple_left}, we observe that this task is much easier for RL, and our agent is able to achieve near-optimal reward within approximately 200k environment steps. Interestingly, the agent also learns a stable strategy of switching off controls once the balls are in the target halves as can be seen in the heatmap visualizations in  Fig.~\ref{fig:orange_right_purple_left}(b) and filmstrip Fig.~\ref{fig:orange_right_purple_left}(c).

\begin{figure}[t!]
    \centering
    \subfloat[][]{\includegraphics[trim={0cm 0cm 1cm 1cm}, clip, width=0.3\linewidth]{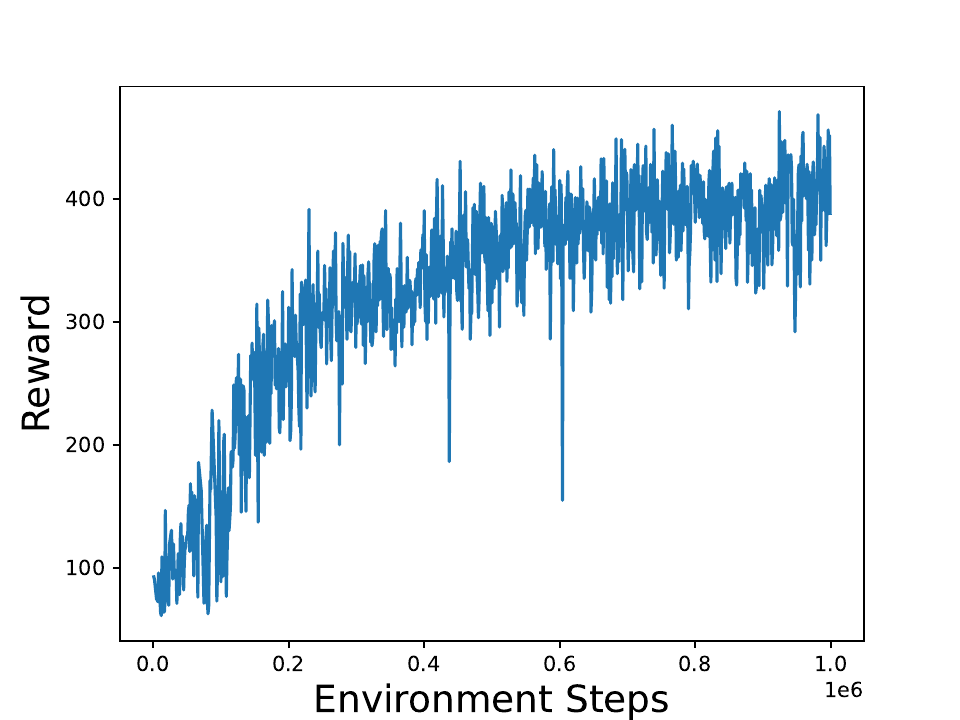}}
    \subfloat[][]{\includegraphics[trim={0cm 0cm 0cm 0cm}, clip, width=0.6\linewidth, height=0.23\linewidth]{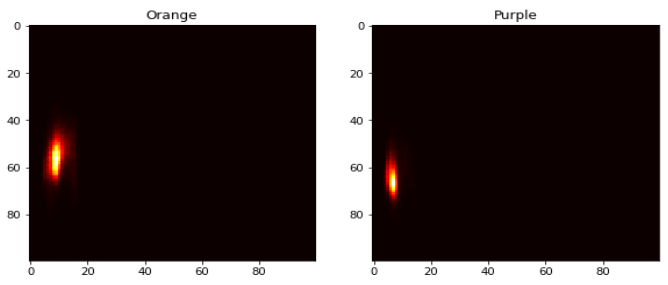}}\\[-0.5ex]
    \subfloat[][]{
    \centering
    \includegraphics[trim={0cm 9cm 0 9cm}, clip, width=0.9\linewidth]{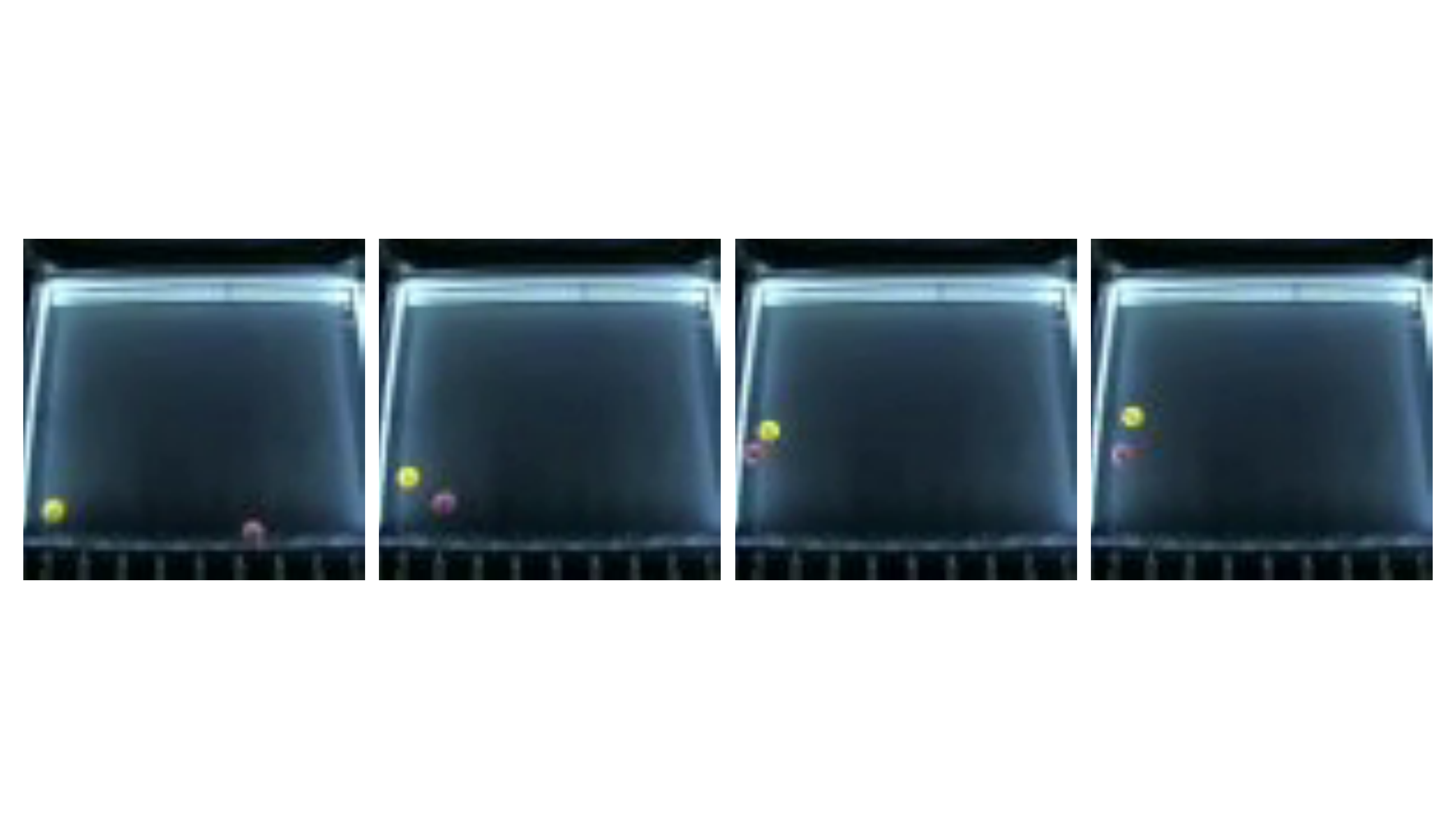}}\hfill
    \caption{Task: Stack orange ball over purple (a) Reward curve. (b) Heatmap visualization of states visited by learned policy (averaged over 100 episodes). (c) Filmstrip of an episode.}
    \label{fig:stacking}
\end{figure}

\paragraph{Stacking:}
To test if our agent can exploit the airflow at a finer level, we consider a more challenging task of stacking two balls on top of each other. We again provide the agent a product of two simple rewards: keep the y-coordinate of the orange over purple by a fixed value and align x-coordinates. We observe that the agent not only learns to successfully stack the balls Fig.~\ref{fig:stacking}(a), but also discovers an interesting strategy to always align them against the left wall of box as it is easier to control airflow near the walls (Fig.~\ref{fig:stacking}(b)).

\subsection{Learning Goal-conditioned Policies to Analyze Reachability}
\begin{figure}[t!]
    \subfloat[][]{\includegraphics[trim={0cm 0cm 0cm 0cm}, width=0.3\linewidth ]{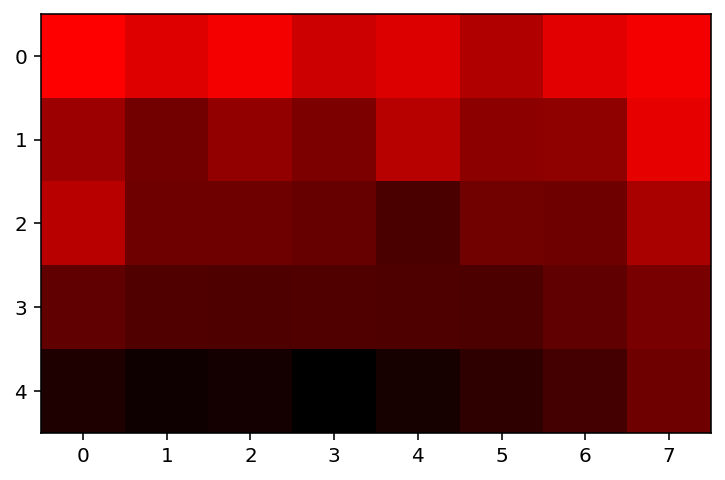}}
    \subfloat[][]{
    \includegraphics[trim={0cm 9cm 0 9cm}, clip, width=0.7\linewidth]{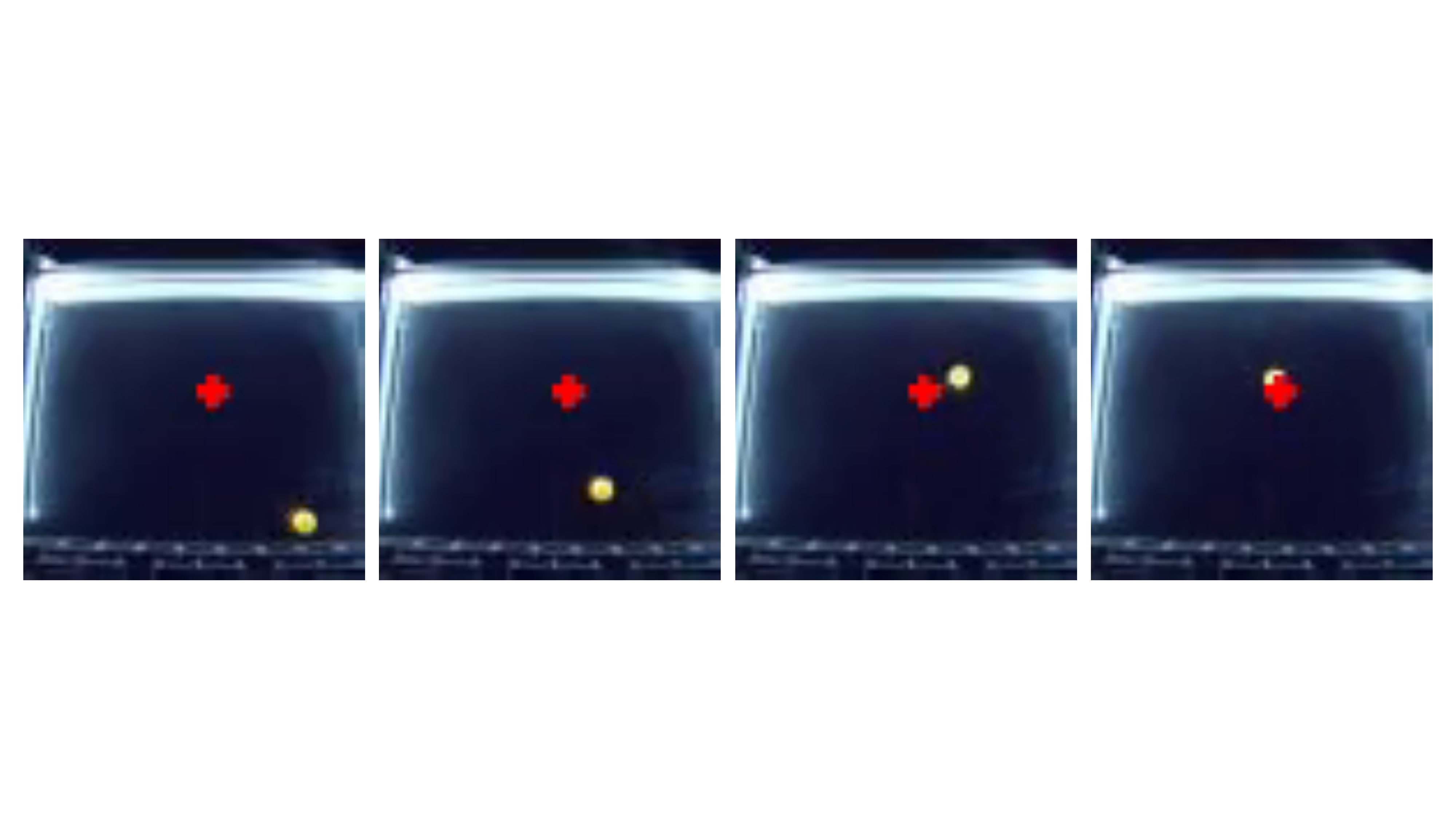}}\hfill
\caption{(a) Pixel intensity is proportional to cumulative error for episodes when the target was in that pixel's bin. Error is the average distance between the ball and target in the last 200 episode steps. (b) Filmstrip of an episode.}
\label{fig:reaching_error_heatmap}
\end{figure}

We wish to characterize what parts of the Box o' Flows are reachable given the actuator configuration and limits. Since, it is not possible analytically, we leverage our RL agent by designing a goal reaching task where the agent must position a ball to randomly chosen pixel targets. We add the goal location to the observation, and train MPO for 1.2M environment steps (1200 episodes). We visually analyze reachability by plotting a coarsely discretized heatmap of reaching errors for different target regions (Fig.~\ref{fig:reaching_error_heatmap}). The intensity of each bin is proportional to the cumulative reaching error for every training episode during which the target was in that bin (normalized such that black is minimum error and red is maximum). This accounts for noise due to policy training and exploration, target height and inherent system stochasticity. The analysis clearly shows that target locations closer to the bottom and center are easier to reach in general. Also, targets near the bottom right are harder than bottom-left and bottom-center, which reveals an imbalance in the airflow through different nozzles. Interestingly, targets closer to the walls are also easily reachable since the agent can better exploit the airflow. These findings also align with the behavior learned in the stacking task. The hardest regions to reach are at the top, especially top-left and top-right corners.

\subsection{Re-using Past Experience via Offline RL}\label{subsec:experiments_offlinerl}
\begin{figure}[t!]
    \centering
    \subfloat[][]{\includegraphics[trim={0cm 0cm 1cm 1cm}, clip, width=0.3\linewidth]{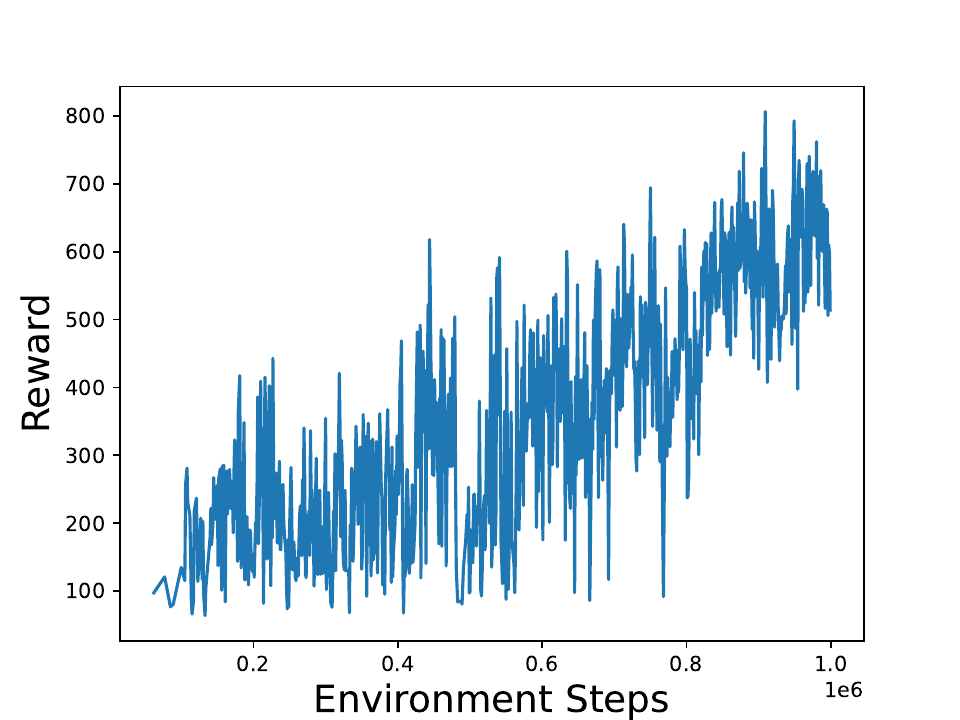}}\hfill
    \subfloat[][]{\includegraphics[trim={0cm 0cm 0cm 0cm}, clip, width=0.7\linewidth]{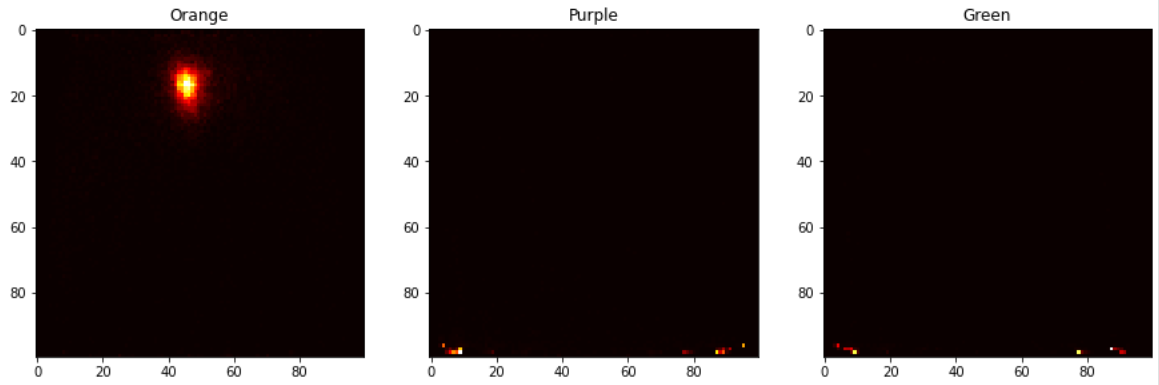}}\\[-0.5ex]
    \subfloat[][]{
    \centering
    \includegraphics[trim={0cm 9cm 0cm 9cm}, clip, width=0.9\linewidth]{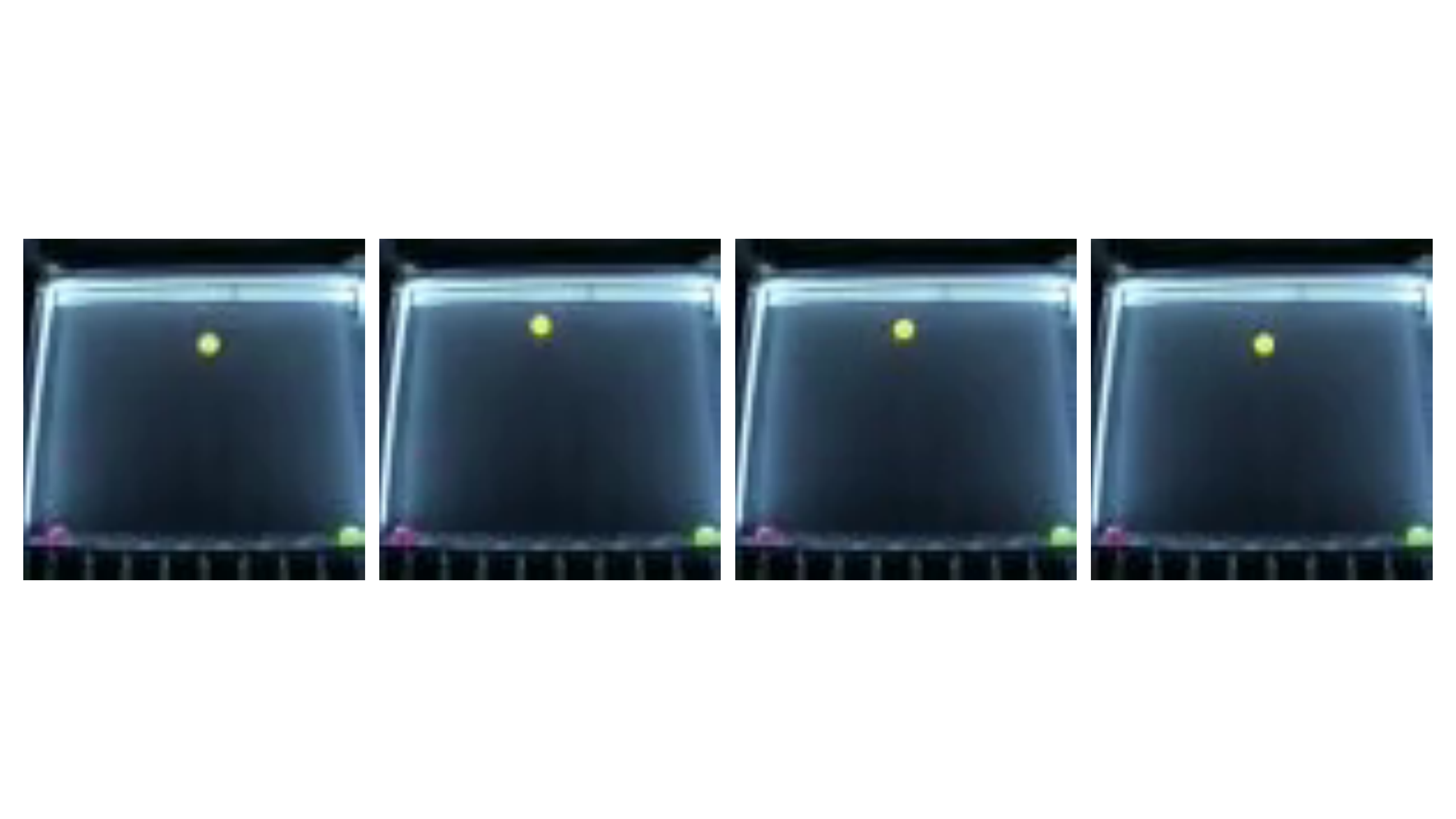}}\hfill
    \centering
    \caption{Task: Maximize the height of orange ball while aligning along the vertical center line in presence of distractors (a) Reward curve and (b) Heatmap visualization of states visited by learned policy (averaged over 100 episodes)(c) Filmstrip of an episode.}
    \label{fig:maximize_orange_center}
\end{figure}

As discussed in Sec.~\ref{sec:methods}, we perform a preliminary experiment to study how offline RL from logged datasets obtained from online RL experiments can be used to test new reward functions. If the logged data has sufficient coverage (i.e the target task is close enough) one can expect the learned policy from offline RL to be representative of what we can obtain by running online RL from scratch. Specifically, we use data from the task of hovering with distractors and re-label the rewards to additionally constrain the ball to remain close to the vertical center line. We then train CRR (Sec.~\ref{sec:methods}) and evaluate the current learner's policy intermittently on the real system. We show the learning curve in Fig.~\ref{fig:maximize_orange_center}(a) and a heatmap of the states visited by the learned policy in Fig~\ref{fig:maximize_orange_center}(b). A stark difference is observed compared to the heatmap in Fig.~\ref{fig:maximize_orange}(b) as the states concentrate entirely near the center as desired, while distractors are at different bottom corners. This experiment provides a promising first result for applying offline RL to study complex dynamical systems like Box o' Flows.  

\section{Related Work}
\paragraph{Deep RL for Complex Physical Systems:}
In addition to real-world robotics discussed in Sec.~\ref{sec:introduction}, RL is also applied to control other complex systems, such as data center cooling systems~\citep{NEURIPS2018_059fdcd9}.~\cite{degrave2022magnetic} apply deep RL to control Tokamak plasmas in nuclear fusion reactors. This is a high dimensional dynamic control problem, however, they rely on simulation in a constrained regime to learn policies that transfer to the real system. \looseness=-1 

\paragraph{Machine Learning for Fluid Dynamics:}
Machine learning and deep RL are being extensively used for the modelling and control of fluid dynamical systems. We provide an overview here and refer the reader to the review papers by ~\citet{brunton2020machine} and \citet{viquerat2022review} for a comprehensive treatment.

\begin{enumerate}[wide, labelwidth=!, itemindent=!, labelindent=0pt]
    \item \textbf{Flow Modelling \& Control}: Machine learning is leveraged to accelerate high-fidelity numerical simulations of fluid dynamics~\citep{doi:10.1073/pnas.2101784118} and automatic turbulence modelling~\citep{novati2021publisher}. Deep RL is also applied to active flow control~\citep{doi:10.1073/pnas.2004939117} and deformable object manipulation~\citep{xu2022dextairity}. The work by~\cite{ma2018fluid} on rigid body manipulation via directed fluid flow is the closest to ours, however, they are limited to simulation with several approximations for computational efficiency.
    \item \textbf{Modelling Biological Systems:} Deep RL can aid the understanding of physical mechanisms and decision-making processes underlying animal behavior.
    ~\cite{verma18swim} combine RL with high-fidelity fluid simulation to study how schooling helps fish reduce energy expenditure. However, running such simulations requires computational resources which are prohibitive for most practitioners. The flight behavior of birds is also studied to design agile UAVs.~\cite{tedrake2009learning} design a glider that demonstrates perching under high angle of attack and~\cite{reddy2016learning} learn energy efficient soaring behaviors by creating numerical models of turbulent thermal convective flows based on bird flight. 
\end{enumerate} 

\paragraph{Offline RL:} 
Offline RL aims to learn competitive policies using logged data without further exploration, and consists of both model-free~\citep{kumar2020conservative, cheng2022adversarially, kostrikov2021offline}, and model-based~\citep{yu2021combo, bhardwaj2023adversarial,kidambi2020morel} variants. A key challenge is offline policy evaluation under limited data coverage~\citep{levine2020offline} which is generally solved by importance sampling based approaches~\citep{precup2000eligibility}. We tackle this via intermittent evaluations of the learner's policy on the real system.

\section{Discussion}
We presented Box o' Flows, a novel benchtop fluid-dynamic control system geared towards real-world RL research. We empirically demonstrated how model-free RL can be used to learn diverse dynamic behaviors directly on hardware, and the applicability of offline RL for efficient re-use of past experience. However, the capabilities of the learning agent can be further enhanced. First, model-based RL methods can be utilized to enhance the understanding of system dynamics and share data among tasks. Second, while our preliminary experiment with offline RL offers promising results, we expect we can improve performance by leveraging methods such as~\cite{cheng2022adversarially} that provide robust policy improvement guarantees. Last but not least, there are many variants of such table top systems that can be realized fairly straightforwardly to vary the difficulty and scope of the experiment.  

\acks{The authors would like to thank IgH for their contribution to the design and engineering of the Box o'Flows and the Google DeepMind London Robotics Lab team for engineering and operational support.}

\bibliography{references}
\end{document}